\documentclass{article}
\PassOptionsToPackage{numbers,sort&compress}{natbib}

\usepackage[sglblindworkshop,final]{neurips_2026}
\workshoptitle{Interpretability as a Science}

\usepackage{amsmath,amssymb,amsthm,mathtools}
\usepackage{booktabs,graphicx,microtype,enumitem}
\usepackage{hyperref,url,xcolor}
\usepackage[most]{tcolorbox}
\usepackage{float}
\definecolor{questionblue}{RGB}{45,90,140}
\graphicspath{{figures/}}
\hypersetup{colorlinks=true,linkcolor=black,citecolor=black,urlcolor=blue!60!black}

\newtheorem{proposition}{Proposition}

\newcommand{\Or}{\mathrm{O}}
\newcommand{\TopK}{\operatorname{TopK}}
\newcommand{\rank}{\operatorname{rank}}

\title{Evidence for Shared Routing Geometry and Dynamics in Sparse Mixture-of-Experts}
\author{
Kirill Labzin \\
Central University, Moscow \\
\texttt{labzin.kr@gmail.com}
\And
Stepan Kulibaba \\
Innopolis University, Innopolis \\
\texttt{kulibabast@gmail.com}
\AND
Artem Dzhalilov \\
Innopolis University, Innopolis \\
\texttt{artem.dzhalilov@gmail.com}
\And
Artem Gorokhov \\
Independent Researcher \\
\texttt{gorohovartem5534@gmail.com}
}

\begin{document}
\maketitle

\begin{abstract}
Sparse mixture-of-experts (MoE) models use an independently parameterized router at each sparse layer to select experts for every token. Prior work has shown that routing decisions across depth can often be predicted from earlier routing signals, suggesting that routing is not fully independent across layers. However, the structure behind this predictability remains unclear. In this work, we provide evidence that routing-relevant states across layers share a common geometric structure that is obscured by layer-specific coordinate systems. We isolate the control subspace of each router and align these spaces into a shared canonical representation using generalized orthogonal Procrustes analysis. After alignment, a single linear transition reaches $R^2=0.39$--$0.71$ and retains 79--90\% of the predictive power of separately fitted layer-specific dynamics, indicating that much of routing-state evolution follows a reusable process across depth. We then ask whether this shared dynamics is specific to routing or simply reflects the smooth evolution of hidden representations. A matched-rank comparison shows that residual representations are often easier to predict across layers, while router-control states preserve the model's expert choices much more faithfully. This separates generic cross-layer predictability from routing-specific information. Finally, we test whether the predicted canonical states remain meaningful when used in place of native routing states. The transported states preserve local routing behavior, while learned state evolution reduces $\Delta\mathrm{NLL}$ relative to simple persistence by 15.7\% on OLMoE and 6.2\% over a 10-router horizon on Phi.
\end{abstract}

\section{Introduction}
Sparse MoE models increase parameter capacity while activating only a subset of experts for each token. Their routers therefore play a central role in determining which computation is performed, how expert load is distributed, and which experts must be available at inference time. Existing work has studied load balancing, specialization, routing failures, and expert prefetching \cite{shazeer2017outrageously,fedus2022switch,zhou2022expertchoice,muennighoff2025olmoe,yoon2026misrouting}. Most analyses, however, treat the router at layer $\ell$ as meaningful only inside that layer's own expert-index space.

Recent evidence suggests a more structured picture. Cross-layer prefetching methods predict future expert choices from nearby routing signals \cite{fang2026fate,zhu2025preattention,zhao2026stmoe}. Mechanistic work identifies a routing-visible component of the residual stream and observes that it rotates across depth \cite{ye2026polysemantic}. Router and expert parameters also develop coupled geometry within a layer \cite{ahrac2026routergeometry}. Together, these observations motivate a basic question:

\begin{tcolorbox}[
    colback=questionblue!5,
    frame hidden,
    arc=2pt,
    left=8pt,
    right=8pt,
    top=6pt,
    bottom=6pt,
    boxrule=0.8pt
]
\centering
What geometric structure underlies the observed predictability of routing across depth, and can routing-relevant states from different layers be aligned into a common representation?
\end{tcolorbox}

Answering this question requires a representation that can be compared meaningfully across layers. Directly comparing router weights is not sufficient, because hidden representations evolve with depth and expert identities are specific to each layer. We therefore isolate, at each layer, the subspace of hidden-state directions that can change the router's relative expert logits. This router-control subspace captures the part of the representation that is functionally visible to the router. We then exploit its orthogonal gauge freedom and use generalized orthogonal Procrustes analysis (GPA) to align the layer-wise control spaces into a common canonical coordinate system.

This common coordinate system allows us to test whether cross-layer routing predictability reflects a reusable dynamical structure rather than only local similarity. If such a structure exists, a single low-capacity transition should predict the evolution of canonical routing states across many layers nearly as well as separately fitted layer-specific transitions. We further test whether these states are functionally meaningful by replacing native routing states with transported ones and measuring the resulting change in language-model loss. If the shared structure captures the routing information actually used by the model, these substitutions should cause little degradation. Our experiments support both predictions, while also showing that the strength of the shared structure depends on the architecture and transport horizon.

\paragraph{Our main contributions are:}
\begin{itemize}[leftmargin=*,topsep=2pt,itemsep=1pt,parsep=0pt]
    \item We show that the router-control factorization has an exact orthogonal gauge symmetry and formulate cross-layer comparison on the corresponding quotient space.
    \item Across Granite, OLMoE, Phi-tiny-MoE, and an IBM shared-expert MoE, one shared linear transition captures 79--90\% of the $R^2$ of layer-specific dynamics after gauge fixing.
    \item Matched-rank, matched-readout probes distinguish generic residual smoothness from information that is specifically useful for expert selection.
    \item Causal interventions show reproducible local transport. A simple margin bound connects canonical-state error to exact top-$k$ stability, and learned dynamics beats persistence at longer horizons on Phi and OLMoE.
\end{itemize}

\section{Related Work}
\paragraph{MoE routing and specialization.} Sparse gating underlies large-scale MoE systems such as sparsely gated MoE and Switch Transformers \cite{shazeer2017outrageously,fedus2022switch}. Later work studies routing schemes and expert specialization in models such as Expert Choice, OLMoE, and OpenMoE \cite{zhou2022expertchoice,muennighoff2025olmoe,openmoe2024}. We instead study how routing-relevant representations relate \emph{across depth}.

\paragraph{Routing structure across depth.} Polysemantic Experts, Monosemantic Paths decomposes hidden states into routing-visible control and routing-invisible content, $h_\ell=h_\ell^{\mathrm{ctrl}}+h_\ell^{\mathrm{content}}$, and observes that control rotates across layers \cite{ye2026polysemantic}. Fate and later prefetching methods likewise show that future expert choices are predictable from earlier routing signals \cite{fang2026fate,zhu2025preattention,zhao2026stmoe}. These results reveal cross-layer routing structure, but not whether it reflects shared dynamics or layer-specific coordinates. We address this by aligning router-control states and studying their shared geometry and dynamics.

\paragraph{Representation alignment.} CKA and Procrustes-based methods provide tools for comparing neural representations beyond their native coordinate systems \cite{kornblith2019,williams2021shape}, while recent work extends orthogonal alignment to multiple representations using GPA \cite{achara2026multiway}. We instead apply this idea to different depths of the same MoE, using the orthogonal non-identifiability of the router-control factorization to define a common coordinate system for routing states.

\paragraph{Layer-wise dynamics.} Viewing network depth as discrete time has precedents in dynamical-systems and Koopman-style analyses \cite{aswani2026koopman}. Rather than modeling the full hidden state, we study router-control states and test whether their apparently layer-specific transitions become shared after alignment.

\section{Method}
\begin{figure}[ht]
\centering
\includegraphics[width=1\columnwidth]{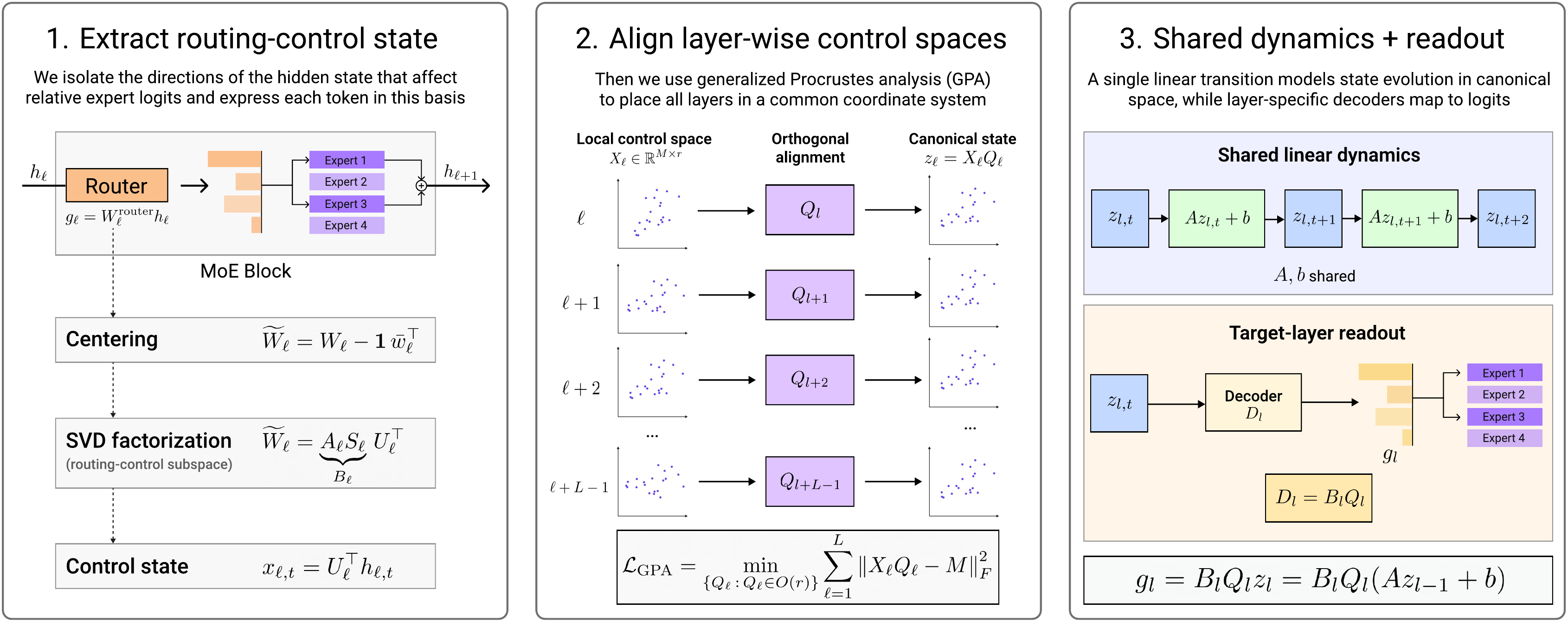}
\caption{
\textbf{Method overview.}
We extract router-control states from layer-wise MoE routers, align them into a shared canonical space, and model their evolution with a shared linear transition. The predicted canonical state is then decoded by the target layer's own readout to recover expert logits and routing decisions.
}
\label{fig:method_overview}
\label{fig:pipeline}
\end{figure}

Figure~\ref{fig:method_overview} summarizes the method in three stages. We first extract a low-dimensional router-control state from the hidden representation entering each sparse MoE router. To compare these layer-specific control states across depth, we align them into a common canonical space using generalized orthogonal Procrustes analysis. In this shared space, we model routing-state evolution with a single linear transition and decode the predicted state with the target layer's own readout to obtain expert logits and routing decisions.

\subsection{Router-control factorization}
At layer $\ell$, let $E$ denote the number of routed experts and $d$ the residual-stream width. The router weight matrix is $W_\ell\in\mathbb{R}^{E\times d}$, the token representation entering the router is $h_{\ell,t}\in\mathbb{R}^{d}$, and the router logits are $g_{\ell,t}=W_\ell h_{\ell,t}\in\mathbb{R}^{E}$. Let $\mathbf 1_E=(1,\ldots,1)^\top\in\mathbb{R}^{E}$ be the all-ones vector and $I_E$ the $E\times E$ identity. Since softmax probabilities and top-$k$ selection are invariant to a common shift of all $E$ expert logits, we remove this common-logit direction by centering the router across experts:
\begin{equation}
\widetilde W_\ell
=\left(I_E-\frac{1}{E}\mathbf 1_E\mathbf 1_E^\top\right)W_\ell
=W_\ell-\mathbf 1_E\bar w_\ell^\top,
\label{eq:centered_router}
\end{equation}
where $\bar w_\ell=E^{-1}W_\ell^\top\mathbf 1_E\in\mathbb{R}^{d}$ is the mean router row. Thus $\widetilde W_\ell$ is simply $W_\ell$ with the mean expert-weight vector subtracted from every row, and $\rank(\widetilde W_\ell)\leq E-1$. Let
\begin{equation}
\widetilde W_\ell=A_\ell S_\ell U_\ell^\top
=B_\ell U_\ell^\top,
\qquad B_\ell\coloneqq A_\ell S_\ell,
\label{eq:router_svd}
\end{equation}
be the compact SVD at rank $r$. The columns of $U_\ell\in\mathbb{R}^{d\times r}$ span exactly the residual-stream directions that can change \emph{relative} router logits. We therefore define the router-control coordinates
\begin{equation}
x_{\ell,t}=U_\ell^\top h_{\ell,t}\in\mathbb{R}^{r},
\qquad
\widetilde g_{\ell,t}=B_\ell x_{\ell,t}.
\label{eq:control_state}
\end{equation}
At full numerical rank, Eq.~\eqref{eq:control_state} reconstructs centered logits to numerical precision. Across all evaluated models, it also preserves the native top-$k$ set exactly and leaves language-model loss unchanged.

\subsection{Exact orthogonal gauge symmetry}
The basis $U_\ell$ is defined only up to an orthogonal change of coordinates. This freedom is not an artifact of the analysis, but an exact symmetry of the router factorization. Let
\[
\Or(r)=\{R\in\mathbb{R}^{r\times r}:R^\top R=I_r\}
\]
denote the set of orthogonal transformations of the $r$-dimensional control space.

\begin{proposition}[Router gauge invariance]
\label{prop:gauge}
For any $R\in\Or(r)$, define
\begin{equation}
U_\ell'=U_\ell R,\qquad
B_\ell'=B_\ell R,\qquad
x_{\ell,t}'=R^\top x_{\ell,t}.
\end{equation}
Then
\begin{equation}
B_\ell'x_{\ell,t}'=B_\ell x_{\ell,t}
\end{equation}
for every token $t$. Thus, an orthogonal change of basis changes the control coordinates but leaves the router output unchanged.
\end{proposition}

Therefore, the particular coordinates $X_\ell$ are not unique: any rotated representation $X_\ell R$ describes the same router-control state. We collect all such representations into the equivalence class
\begin{equation}
[X_\ell]=\{X_\ell R:R\in\Or(r)\}.
\end{equation}
Cross-layer comparison should therefore ignore this arbitrary choice of basis. A natural way to do so is the orthogonal Procrustes distance \cite{williams2021shape},
\begin{equation}
d_{\mathrm P}([X],[Y])
=
\min_{R\in\Or(r)}\|XR-Y\|_F,
\label{eq:proc_distance}
\end{equation}
which first finds the best orthogonal alignment and then measures the remaining discrepancy. This optimization has the closed form
\begin{equation}
d_{\mathrm P}^2
=
\|X\|_F^2+\|Y\|_F^2-2\|X^\top Y\|_*.
\label{eq:proc_closed}
\end{equation}

Thus, Procrustes alignment resolves the orthogonal basis ambiguity inherent in the router-control factorization and enables meaningful cross-layer comparison.

\subsection{A globally consistent canonical gauge}
Pairwise Procrustes fits need not define a single coordinate system across depth. We therefore use generalized Procrustes analysis (GPA). On training tokens, after layer-wise centering and scalar normalization, we solve
\begin{equation}
\min_{\{Q_\ell\},M}
\sum_{\ell=1}^{L}\|X_\ell Q_\ell-M\|_F^2,
\qquad Q_\ell\in\Or(r).
\label{eq:gpa}
\end{equation}
The canonical state of a column vector $x_{\ell,t}$ is then
\begin{equation}
z_{\ell,t}=Q_\ell^\top\,\mathrm{norm}(x_{\ell,t}).
\label{eq:canonical_state}
\end{equation}
For fixed aligned clouds $Z_\ell=X_\ell Q_\ell$, the optimal template is their mean $M^*=L^{-1}\sum_\ell Z_\ell$. Moreover,
\begin{equation}
\sum_\ell\|Z_\ell-M^*\|_F^2
=\frac{1}{L}\sum_{\ell<m}\|Z_\ell-Z_m\|_F^2,
\label{eq:barycenter_identity}
\end{equation}
so GPA can equivalently be viewed as minimizing total pairwise disagreement among layer gauges. All $Q_\ell$ are fitted on training sequences only.

\subsection{Shared dynamics as gauge reduction}
Suppose raw control coordinates admit layer-specific affine transitions. Ignoring the fixed normalization for notational clarity,
\begin{equation}
x_{\ell+1}\approx F_\ell x_\ell+c_\ell,
\qquad
z_\ell=Q_\ell^\top x_\ell
\;\Longrightarrow\;
x_\ell=Q_\ell z_\ell,
\label{eq:raw_transition}
\end{equation}
where the last relation follows from the orthogonality of $Q_\ell$.

After the canonicalization in Eq.~\eqref{eq:canonical_state}, the corresponding transition becomes
\begin{align}
z_{\ell+1}
&= Q_{\ell+1}^\top x_{\ell+1} \\
&\approx Q_{\ell+1}^\top (F_\ell x_\ell + c_\ell) \\
&= Q_{\ell+1}^\top F_\ell Q_\ell z_\ell
   + Q_{\ell+1}^\top c_\ell.
\label{eq:gauge_reduced_transition}
\end{align}
Our central dynamical hypothesis is therefore precise: there exists a low-capacity operator $A$ such that
\begin{equation}
Q_{\ell+1}^\top F_\ell Q_\ell\approx A
\quad\text{for many layers }\ell,
\label{eq:shared_operator}
\end{equation}
or equivalently $F_\ell\approx Q_{\ell+1}A Q_\ell^\top$. Empirically, we fit
\begin{equation}
\hat z_{\ell+1,t}=Az_{\ell,t}+b
\label{eq:shared}
\end{equation}
by ridge regression on pooled layer transitions, with regularization selected on validation data. Layer-specific affine models provide an upper bound. We summarize the degree of reuse with
\begin{equation}
\eta_{\mathrm{share}}
=\frac{R^2_{\mathrm{shared}}}{R^2_{\mathrm{layer\text{-}specific}}},
\label{eq:sharing_ratio}
\end{equation}
which equals one when a single transition is as predictive as fitting each layer independently.

\subsection{Routing stability under transport}

To evaluate a predicted canonical state, we first map it back to the expert logits of the target layer. Ignoring the fixed normalization transform, which can be absorbed into the decoder and a bias, we have $x_\ell=Q_\ell z_\ell$ and therefore
\begin{equation}
\widetilde g_{\ell,t}
=
B_\ell x_{\ell,t}
=
B_\ell Q_\ell z_{\ell,t}
=
D_\ell z_{\ell,t},
\qquad
D_\ell:=B_\ell Q_\ell.
\label{eq:canonical_decoder}
\end{equation}
Thus, $D_\ell$ is the layer-specific decoder from canonical routing coordinates to centered expert logits.

Now let $\hat z=z+e$ be a transported or predicted canonical state, where $e$ is its state-space error. Its decoded logits satisfy
\begin{equation}
D_\ell\hat z
=
D_\ell z + D_\ell e,
\label{eq:logit_error}
\end{equation}
so $D_\ell e$ is exactly the induced error in the centered logits. Let
\[
\widetilde g_{(1)}\geq\widetilde g_{(2)}\geq\cdots\geq\widetilde g_{(E)}
\]
denote the sorted true logits, and define the top-$k$ routing margin
\begin{equation}
\gamma_k
=
\widetilde g_{(k)}-\widetilde g_{(k+1)}.
\label{eq:routing_margin}
\end{equation}

\begin{proposition}[Top-$k$ stability]
\label{prop:topk}
If
\begin{equation}
\|D_\ell e\|_\infty
<
\frac{\gamma_k}{2},
\label{eq:topk_bound}
\end{equation}
then the selected expert set is unchanged:
\begin{equation}
\TopK(D_\ell\hat z)
=
\TopK(D_\ell z).
\end{equation}
A sufficient condition directly in canonical state space is
\begin{equation}
\|e\|_2
<
\frac{\gamma_k}
{2\|D_\ell\|_{2\rightarrow\infty}},
\qquad
\|D_\ell\|_{2\rightarrow\infty}
=
\max_{j\in[E]}\|d_{\ell,j}\|_2.
\label{eq:state_margin_bound}
\end{equation}
\end{proposition}

Proposition~\ref{prop:topk} connects canonical-state prediction error to the discrete routing decision. The same state error can leave routing unchanged when the top-$k$ margin is large, but change the selected experts for a token close to the routing boundary. Geometric prediction quality alone therefore does not determine routing stability. Proofs of Propositions~\ref{prop:gauge} and~\ref{prop:topk}, together with the GPA identity, are given in Appendix~\ref{app:math}.

\section{Experimental Setup}

We evaluate four sparse MoE architectures chosen to vary depth, number of experts,
routing sparsity, and the presence of explicit shared experts
(Table~\ref{tab:models}).

\vspace{-2pt}
\begin{table}[H]
\centering
\caption{Evaluated MoE architectures. Rank is the numerical rank of the centered router.}
\label{tab:models}
\small
\setlength{\tabcolsep}{4pt}
\renewcommand{\arraystretch}{0.95}
\begin{tabular}{lrrrr}
\toprule
Model & Layers & Experts & Top-$k$ & Rank\\
\midrule
Granite    & 24 & 32 & 8 & 31\\
OLMoE-SFT  & 16 & 64 & 8 & 63\\
Phi-tiny   & 32 & 16 & 2 & 15\\
IBM Shared & 40 & 62 & 6 & 61\\
\bottomrule
\end{tabular}
\end{table}

\paragraph{Data and splits.}
We use WikiText-2 raw text as a common natural-language probe corpus. Geometry, alignment baselines, and dynamics ablations use three independent sequence-level splits (seeds 101, 202, 303), each with 16,384 held-out tokens. Final matched-readout validation and targeted causal replications use 50,176 held-out tokens per seed (392 sequences of length 128). Normalization, GPA, dynamics, and readout fitting use training data only, while hyperparameters are selected on validation data.

\paragraph{Metrics.}
We evaluate geometry using held-out same-token cosine similarity and linear CKA. Dynamics are measured with $R^2$, normalized MSE, and cosine similarity. Routing fidelity is measured with top-1 accuracy, top-$k$ recall and Jaccard, Jensen--Shannon divergence, and centered-logit MSE. We evaluate causal interventions using $\Delta$NLL.

\section{Results}
We present the results in five parts. First, we test whether aligning router-control states across layers makes their evolution more predictable with a single shared transition. Second, we compare router-control states with the residual stream to determine whether this predictability is specific to routing or simply reflects general similarity between nearby layers. Third, we study how the dimensionality of the representation affects predictability and routing accuracy. Fourth, we test whether the aligned states can replace the original routing states without strongly affecting model performance. Finally, we examine where this shared structure breaks down and clarify the limits of our claim.

\subsection{Gauge alignment exposes reusable dynamics}
\begin{figure}[h]
\centering
\begin{minipage}[t]{0.49\linewidth}
\centering
\includegraphics[width=\linewidth]{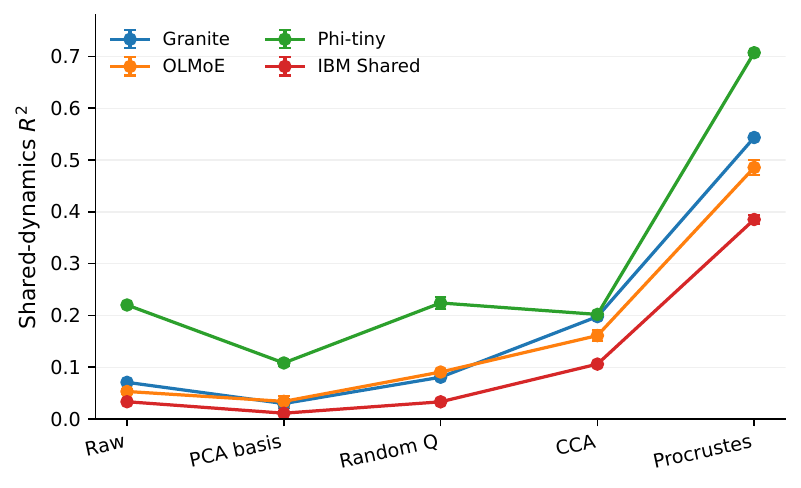}\\[-2pt]
\footnotesize\textbf{(a) Alignment baselines}
\end{minipage}\hfill
\begin{minipage}[t]{0.49\linewidth}
\centering
\includegraphics[width=\linewidth]{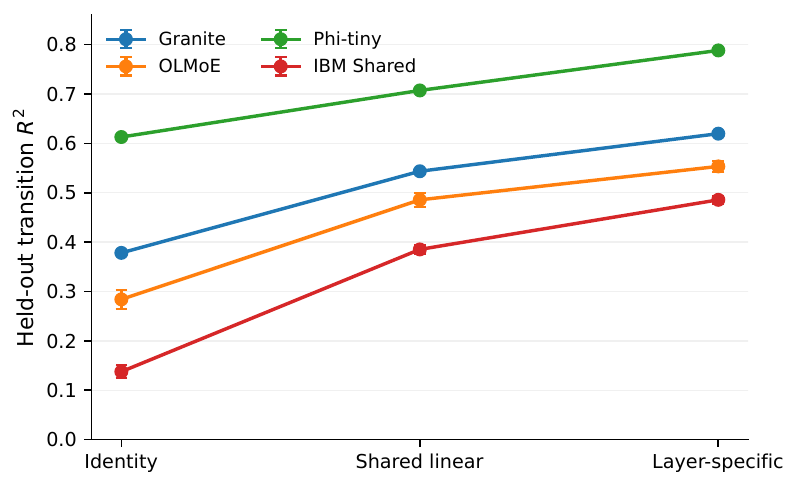}\\[-2pt]
\footnotesize\textbf{(b) Shared vs. layer-specific dynamics}
\end{minipage}
\caption{\textbf{Gauge fixing exposes reusable cross-layer dynamics.} (a) Shared prediction remains weak under raw, PCA-basis, random-gauge, and CCA coordinates, then rises sharply after orthogonal Procrustes alignment. (b) One learned transition closes most of the gap from persistence to separately fitted layer-specific dynamics. Error bars show standard deviation across three split seeds.}
\label{fig:alignment}
\label{fig:shared}
\end{figure}

\textbf{A shared transition is weak in native coordinates and strong after gauge fixing.}
Across architectures, the pooled linear transition remains weak in raw coordinates and under generic alignment baselines, but becomes substantially more predictive after orthogonal Procrustes alignment (Fig.~\ref{fig:alignment}). Random orthogonal gauges and PCA-basis alignment do not produce the same improvement, and shuffling the correspondence between tokens removes the alignment effect almost entirely (full ablation in Table~\ref{tab:app_align}). This suggests that the improvement comes from aligning meaningful cross-layer structure rather than simply applying an additional orthogonal transformation.

\textbf{One shared transition explains most of the predictable cross-layer change.}
We next compare the shared transition with a stronger baseline that fits a separate transition for every pair of adjacent layers. As shown in Table~\ref{tab:dynamics_main}, the shared model reaches 79--90\% of the $R^2$ obtained by these layer-specific models across all four architectures, while using far fewer parameters. This shows that the predictable evolution of router-control states is not strongly specific to individual layers. After alignment, much of this evolution can instead be captured by the same transition across depth.

\begin{table}[h]
\centering
\caption{\textbf{Cross-model geometry and dynamics.} Values are means over three split seeds. ``Id.'' is canonical-state persistence, ``Shared'' is one pooled linear transition, and ``Layer'' fits a separate linear transition per depth. $\eta_{\mathrm{share}}=R^2_{\mathrm{Shared}}/R^2_{\mathrm{Layer}}$. Params are shared/layer-specific transition parameters.}
\label{tab:dynamics_main}
\scriptsize
\setlength{\tabcolsep}{2.6pt}
\begin{tabular*}{\linewidth}{@{\extracolsep{\fill}}lrrrrrc}
\toprule
Model & GPA cos. & Id. $R^2$ & Shared $R^2$ & Layer $R^2$ & $\eta_{\mathrm{share}}$ & Params S/L \\
\midrule
Granite    & 0.581 & 0.378 & \textbf{0.544} & 0.620 & 0.88 & 0.99k / 22.8k \\
OLMoE      & 0.548 & 0.284 & \textbf{0.486} & 0.553 & 0.88 & 4.03k / 60.5k \\
Phi-tiny   & 0.534 & 0.613 & \textbf{0.707} & 0.788 & 0.90 & 0.24k / 7.44k \\
IBM Shared & 0.431 & 0.138 & \textbf{0.385} & 0.486 & 0.79 & 3.78k / 147.5k \\
\bottomrule
\end{tabular*}
\end{table}

\subsection{Residual smoothness is not the same as routing specificity}
A possible alternative explanation is that canonical router states are predictable simply because hidden representations change smoothly across nearby layers. To test this, we compare them with matched-rank PCA and random residual subspaces. For each representation, we train a decoder with the same parameter budget to reconstruct the true router logits on held-out tokens.

Table~\ref{tab:readout_main} shows a clear difference. Residual PCA is much easier to predict across layers, while router-control states recover the actual expert choices much more accurately. This shows that cross-layer predictability and routing relevance are different properties: residual representations evolve more smoothly, but router-control states retain the information that is directly used for expert selection.

\begin{table}[h]
\centering
\caption{\textbf{Matched-rank, equal-budget readout comparison.} Means over three 50,176-token test splits. Higher is better for both metrics.}
\label{tab:readout_main}
\small
\setlength{\tabcolsep}{3.6pt}
\begin{tabular*}{\linewidth}{@{\extracolsep{\fill}}lrrrr}
\toprule
& \multicolumn{2}{c}{Next-state $R^2$} & \multicolumn{2}{c}{Top-$k$ recall}\\
\cmidrule(lr){2-3}\cmidrule(lr){4-5}
Model & Router & PCA & Router & PCA\\
\midrule
Granite    & 0.542 & \textbf{0.855} & \textbf{0.999} & 0.798\\
OLMoE      & 0.500 & \textbf{0.788} & \textbf{0.998} & 0.669\\
Phi-tiny   & 0.424 & \textbf{0.861} & \textbf{0.989} & 0.405\\
IBM Shared & 0.371 & \textbf{0.806} & \textbf{0.998} & 0.589\\
\bottomrule
\end{tabular*}
\end{table}

\subsection{A low-dimensional shared core coexists with routing detail}
The rank sweep shows that the most predictable representation is not necessarily the one that best preserves routing behavior. OLMoE provides the clearest example (Table~\ref{tab:olmoe_rank}): very low-rank states are highly predictable, while higher ranks are needed to recover expert choices accurately and to improve causal transport. This suggests that a small shared component captures much of the cross-layer dynamics, while additional dimensions preserve finer routing information.

\begin{table}[h]
\centering
\caption{\textbf{OLMoE rank trade-off} on the screening split (seed 101). Lower causal $\Delta$NLL and higher $R^2$/recall are better. The full centered-router rank is 63.}
\label{tab:olmoe_rank}
\small
\setlength{\tabcolsep}{4.3pt}
\begin{tabular*}{\linewidth}{@{\extracolsep{\fill}}rrrr}
\toprule
Rank & Shared $R^2$ & Top-$k$ recall & Causal $\Delta$NLL\\
\midrule
2  & \textbf{0.695} & 0.316 & 0.593\\
8  & 0.614 & 0.492 & 0.309\\
16 & 0.585 & 0.568 & 0.230\\
31 & 0.541 & 0.638 & 0.176\\
63 & 0.487 & \textbf{0.699} & \textbf{0.118}\\
\bottomrule
\end{tabular*}
\end{table}

\begin{figure}[h]
\centering
\begin{minipage}[t]{0.49\linewidth}\centering
\includegraphics[width=\linewidth]{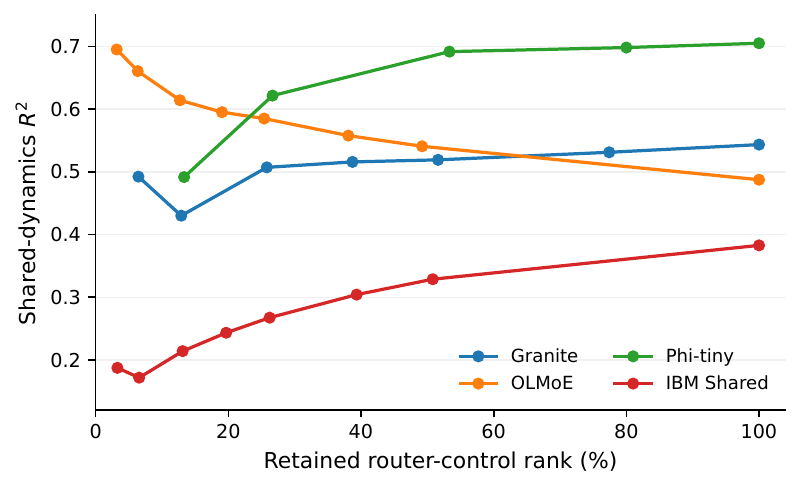}\\[-2pt]
\footnotesize\textbf{(a) Shared-dynamics $R^2$}
\end{minipage}\hfill
\begin{minipage}[t]{0.49\linewidth}\centering
\includegraphics[width=\linewidth]{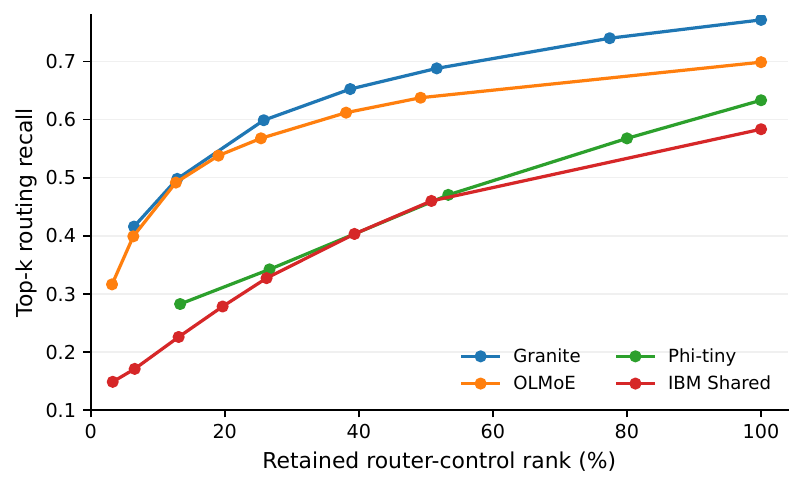}\\[-2pt]
\footnotesize\textbf{(b) Top-$k$ routing recall}
\end{minipage}
\caption{\textbf{A low-dimensional shared structure captures much of the predictable dynamics, while accurate routing requires more dimensions.} OLMoE shows this most clearly: low-rank states are easy to predict, but higher ranks are needed to recover expert choices accurately. Rank is normalized by each model's full centered-router rank.}
\label{fig:rank_sweep}
\end{figure}

In centered coordinates, the unregularized least-squares shared operator is $A^*=C_{10}C_{00}^{\dagger}$, where $C_{10}=\mathbb E[z_{\ell+1}z_\ell^\top]$ and $C_{00}=\mathbb E[z_\ell z_\ell^\top]$. The rank sweep therefore measures how many control dimensions are needed to preserve predictable cross-layer structure and how many are needed for accurate routing decisions. The IBM shared-expert model is less compressible: both predictability and routing fidelity continue to improve as rank increases (Fig.~\ref{fig:rank_sweep}).

\subsection{Causal transport tests functional relevance}
Good prediction alone does not show that the aligned states are actually important for model behavior. We therefore replace the native routing states at selected layers with transported states and measure the resulting change in $\Delta$NLL.

\textbf{Canonicalization is important for causal transport.}
For Granite layers 9--13, the shared transition causes much less degradation in canonical coordinates than in raw coordinates, with shuffled gauges, or in a random matched-rank subspace (Table~\ref{tab:granite_causal_controls}). Results are averaged over three 50,176-token splits.

\begin{table}[ht]
\centering
\caption{\textbf{Granite causal controls, layers 9--13.} Mean $\Delta$NLL $\pm$ standard deviation across seeds 101/202/303. Lower is better.}
\label{tab:granite_causal_controls}
\small
\setlength{\tabcolsep}{5pt}
\begin{tabular*}{\linewidth}{@{\extracolsep{\fill}}lr}
\toprule
Transport variant & $\Delta$NLL\\
\midrule
Canonical + shared $A$ & \textbf{$0.0070\pm0.0016$}\\
Canonical + identity   & $0.0142\pm0.0032$\\
Raw coordinates + shared $A$ & $0.1120\pm0.0046$\\
Shuffled $Q$ + shared $A$ & $0.1012\pm0.0060$\\
Random subspace + shared $A$ & $0.1495\pm0.0099$\\
\bottomrule
\end{tabular*}
\end{table}

\textbf{Learned dynamics outperforms persistence across replications.}
Table~\ref{tab} shows consistent gains for OLMoE layers 8--9 and both long Phi horizons. Degradation increases with transport length, while the advantage over persistence becomes more pronounced (Appendix~\ref{app}). The substantial Phi degradation suggests systematic router dynamics, but not a practical router-skipping method.

\begin{table}[h]
\centering
\caption{\textbf{Targeted causal replications.} Means over three fixed 50,176-token splits. The final column is Shared minus Identity; negative values favor learned dynamics.}
\label{tab:causal_rep_main}
\small
\setlength{\tabcolsep}{3.8pt}
\begin{tabular*}{\linewidth}{@{\extracolsep{\fill}}llrrr}
\toprule
Model & Block & Identity & Shared & Shared$-$Id.\\
\midrule
OLMoE    & 8--9   & 0.0375 & \textbf{0.0316} & $-0.0059$\\
Phi-tiny & 24--31 & 0.2670 & \textbf{0.2520} & $-0.0150$\\
Phi-tiny & 22--31 & 0.3217 & \textbf{0.3017} & $-0.0200$\\
\bottomrule
\end{tabular*}
\end{table}

The IBM shared-expert model provides an important counterexample to over-generalization. It still exhibits canonical geometry and shared predictability, and selected short blocks can be transported without measurable degradation in replicated screening tests, but learned $A$ is not uniformly better than identity across horizons. The benefit of explicitly evolving the canonical state is therefore architecture- and horizon-dependent.

\subsection{Negative results define the limits}
The negative results show where the shared structure stops being useful. First, directly averaging or sharing router weights strongly degrades model quality, meaning that aligned control states do not make expert identities or router readouts interchangeable across layers. Second, repeatedly skipping routers causes errors to accumulate quickly. We therefore limit our claim to \emph{local functional transport and reusable cross-layer dynamics}.

\section{Discussion}
The experiments support a three-level picture. The residual state $h_\ell$ contains a smooth cross-layer backbone. The control state $x_\ell=U_\ell^\top h_\ell$ isolates directions that can alter relative expert logits. Orthogonal gauge fixing maps these layer-specific coordinates to $z_\ell$, where a low-capacity transition becomes reusable across depth.

The matched-readout result is central to this interpretation. If our observation were only generic residual smoothness, residual PCA should be an equally good routing representation. It is not. PCA predicts future residual state more accurately, while router-control coordinates preserve expert choices much more faithfully. This separates two properties that are easy to conflate: \emph{temporal predictability} and \emph{routing relevance}.

Our results also make the phrase ``shared dynamics'' more precise. We do not find a universal transition that makes all layers dynamically identical. Rather, Eq.~\eqref{eq:shared_operator} says that apparently different raw transitions are approximately related by layer-specific gauge changes, and one shared transition captures most of the predictive power of independently fitted layer-specific transitions with far fewer parameters. Its causal advantage over identity is clearest at longer horizons on Phi and OLMoE; the explicit shared-expert architecture exhibits weaker universality. The evidence therefore supports reusable cross-layer structure, not exact dynamical equivalence. Proposition~\ref{prop:topk} further separates representation error from routing error: functional stability depends on the transported-state error measured through the logit decoder relative to the local top-$k$ margin.

Finally, the findings complement cross-layer prefetching work. Prefetching systems exploit future-expert predictability operationally \cite{fang2026fate,zhu2025preattention,zhao2026stmoe}; our experiments provide a geometric coordinate system in which a simple shared predictor becomes effective. Conversely, the strong residual-PCA baseline shows that cross-layer predictability is broader than the router subspace itself.

\section{Limitations}
The study is empirical and uses one common text corpus for controlled cross-model comparison. Phi-tiny-MoE is instruction-tuned and therefore distribution-mismatched to WikiText perplexity; we use its causal results to compare interventions rather than claim improved language modeling. The IBM shared-expert checkpoint is a research checkpoint rather than a widely deployed production model. The exact symmetry identifies an equivalence class rather than a unique latent coordinate system: GPA fixes one convenient global gauge, but the canonical state remains globally identifiable only up to a common orthogonal transform. Finally, causal transport is used as an analysis tool: recursive open-loop skipping accumulates error, and we make no wall-clock speedup claim.

\section{Conclusion}
Across four sparse MoE architectures, we find that router-control states contain consistent cross-layer structure that becomes visible after orthogonal gauge alignment. In the canonical space, a single linear transition captures most of the predictive power of layer-specific models while using far fewer parameters. This effect is not explained by generic hidden-state smoothness: residual states are often easier to predict, but router-control states preserve expert choices much more accurately. Causal interventions further show that these aligned states remain functionally meaningful, with learned evolution outperforming persistence at longer horizons on Phi and OLMoE. Together, these results suggest that MoE routers share routing-relevant structure and partially reusable dynamics across depth.

\bibliographystyle{plainnat}
\bibliography{references}

\appendix
\section{Mathematical Details}
\label{app:math}

\paragraph{Proof of Proposition~\ref{prop:gauge}.}
For $R\in\Or(r)$,
\begin{equation}
B_\ell'x_\ell'
=(B_\ell R)(R^\top x_\ell)
=B_\ell(RR^\top)x_\ell
=B_\ell x_\ell.
\end{equation}
Thus all centered logits, softmax probabilities, and top-$k$ decisions are identical. The same argument applies token-wise to a data matrix $X_\ell$, whose coordinate representation transforms as $X_\ell\mapsto X_\ell R$.

\paragraph{Procrustes closed form.}
For centered $X,Y$,
\begin{align}
\min_{R\in\Or(r)}\|XR-Y\|_F^2
&=\|X\|_F^2+\|Y\|_F^2
-2\max_{R\in\Or(r)}\operatorname{tr}(R^\top X^\top Y)\nonumber\\
&=\|X\|_F^2+\|Y\|_F^2-2\|X^\top Y\|_*,
\end{align}
where the maximum is attained by the standard orthogonal Procrustes solution obtained from the SVD of $X^\top Y$.

\paragraph{GPA pairwise-dispersion identity.}
Let $M^*=L^{-1}\sum_\ell Z_\ell$. Expanding squared distances gives
\begin{equation}
\sum_\ell\|Z_\ell-M^*\|_F^2
=\sum_\ell\|Z_\ell\|_F^2-L\|M^*\|_F^2
=\frac{1}{L}\sum_{\ell<m}\|Z_\ell-Z_m\|_F^2.
\end{equation}
Hence minimizing the GPA objective minimizes the average pairwise disagreement among all aligned layer representations.

\paragraph{Proof of Proposition~\ref{prop:topk}.}
Let $i$ be any selected expert and $j$ any unselected expert. By definition of the top-$k$ margin, $g_i-g_j\geq\gamma_k$. With logit perturbation $\delta g=D_\ell e$,
\begin{equation}
\hat g_i-\hat g_j
\geq\gamma_k-|\delta g_i|-|\delta g_j|
>0
\end{equation}
whenever $\|\delta g\|_\infty<\gamma_k/2$. Therefore no selected expert can cross an unselected expert, so the top-$k$ set is unchanged. Finally, $\|D_\ell e\|_\infty\leq\|D_\ell\|_{2\rightarrow\infty}\|e\|_2$ yields Eq.~\eqref{eq:state_margin_bound}.

\paragraph{Least-squares shared operator.}
Ignoring the intercept after centering, the pooled objective is $\mathbb E\|z_{\ell+1}-Az_\ell\|_2^2$. The minimum-norm solution is
\begin{equation}
A^*=C_{10}C_{00}^{\dagger},
\qquad
C_{10}=\mathbb E[z_{\ell+1}z_\ell^\top],\quad
C_{00}=\mathbb E[z_\ell z_\ell^\top].
\end{equation}
This makes explicit that the shared dynamics are determined by cross-layer covariance expressed in the canonical gauge.

\section{Additional Experimental Details}
\begin{figure}[h]
\centering
\begin{minipage}[t]{0.49\linewidth}\centering
\includegraphics[width=\linewidth]{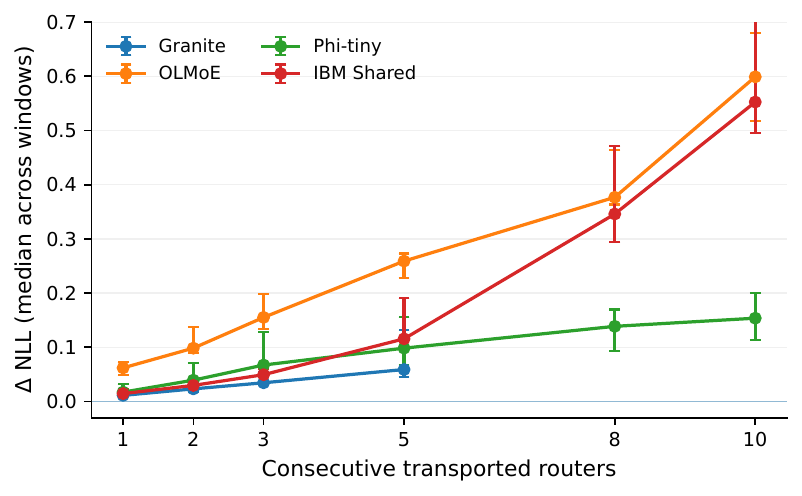}\\[-2pt]
\footnotesize\textbf{(a) Transport horizon vs. $\Delta$NLL}
\end{minipage}\hfill
\begin{minipage}[t]{0.49\linewidth}\centering
\includegraphics[width=\linewidth]{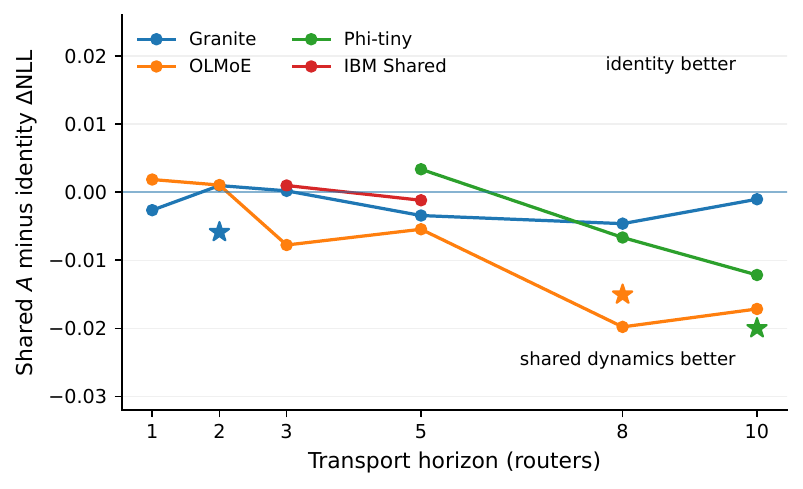}\\[-2pt]
\footnotesize\textbf{(b) Learned dynamics vs. persistence}
\end{minipage}
\caption{\textbf{Causal transport has a clear horizon structure.}
(a) Median $\Delta$NLL over screened contiguous windows with interquartile ranges.
(b) Shared $A$ minus identity $\Delta$NLL; values below zero favor learned dynamics,
and stars mark independently replicated three-seed points.}
\label{fig:causal_horizon}
\end{figure}

\paragraph{Seed-level targeted causal replications.}
\label{app:causal_horizon}
To characterize how transport degrades with distance, we sweep the number of consecutive transported routers over screened contiguous windows. Figure~\ref{fig:causal_horizon} shows two complementary effects. Absolute $\Delta$NLL generally increases with transport horizon, indicating error accumulation under longer interventions. At the same time, the learned shared transition becomes increasingly competitive with identity persistence at longer horizons, particularly for OLMoE and Phi-tiny. Stars mark the fixed three-seed replications reported in Table~\ref{tab:causal_rep_main}.

\paragraph{Reconstruction checks.}
Before every causal run, centered router reconstruction is verified at full numerical rank. Across all layers and models, reconstructed top-$k$ sets agree exactly with native routing. Evaluation scripts also verify model-weight checksums before and after interventions.

\paragraph{Complete alignment controls.}
Table~\ref{tab:app_align} reports the alignment baselines used in Fig.~\ref{fig:alignment}. Values are mean shared-dynamics $R^2$ across the three split seeds. ``Random $Q$'' applies independently sampled orthogonal gauges; ``PCA basis'' aligns via PCA coordinate conventions rather than token-wise Procrustes.

\begin{table}[H]
\centering
\caption{\textbf{Complete alignment ablation.} Mean shared-dynamics $R^2$ across three split seeds.}
\label{tab:app_align}
\small
\setlength{\tabcolsep}{3.5pt}
\begin{tabular*}{\linewidth}{@{\extracolsep{\fill}}lrrrrr}
\toprule
Model & Raw & PCA basis & Random $Q$ & CCA & Procrustes\\
\midrule
Granite    & 0.071 & 0.030 & 0.081 & 0.198 & \textbf{0.544}\\
OLMoE      & 0.053 & 0.034 & 0.091 & 0.161 & \textbf{0.486}\\
Phi-tiny   & 0.220 & 0.108 & 0.224 & 0.202 & \textbf{0.707}\\
IBM Shared & 0.033 & 0.011 & 0.033 & 0.106 & \textbf{0.385}\\
\bottomrule
\end{tabular*}
\end{table}

\paragraph{Matched readout.}
Router-control and residual-PCA representations use learned linear logit decoders with the same per-layer parameterization $(r+1)E$. Exact cross-model means are reported in Table~\ref{tab:readout_main}; random matched-rank residual subspaces are additionally used as a negative control in the experiment code.

\paragraph{Seed-level targeted causal replications.}
Table~\ref{tab:app_causal_seed} expands Table~\ref{tab:causal_rep_main}. The blocks are fixed before evaluation on the additional split seeds; they are not re-selected per seed.

\begin{table}[h]
\centering
\caption{\textbf{Seed-level targeted causal replications.} Lower $\Delta$NLL is better; Shared$-$Id. $<0$ favors learned dynamics.}
\label{tab:app_causal_seed}
\scriptsize
\setlength{\tabcolsep}{3.2pt}
\begin{tabular*}{\linewidth}{@{\extracolsep{\fill}}llrrrr}
\toprule
Model & Block & Seed & Identity & Shared & Shared$-$Id.\\
\midrule
OLMoE & 8--9 & 101 & 0.0386 & 0.0326 & $-0.0060$\\
      &      & 202 & 0.0457 & 0.0418 & $-0.0039$\\
      &      & 303 & 0.0281 & 0.0204 & $-0.0077$\\
\midrule
Phi-tiny & 24--31 & 101 & 0.2552 & 0.2471 & $-0.0080$\\
         &        & 202 & 0.2819 & 0.2654 & $-0.0166$\\
         &        & 303 & 0.2639 & 0.2434 & $-0.0206$\\
\midrule
Phi-tiny & 22--31 & 101 & 0.3112 & 0.3001 & $-0.0110$\\
         &        & 202 & 0.3378 & 0.3120 & $-0.0258$\\
         &        & 303 & 0.3161 & 0.2930 & $-0.0231$\\
\bottomrule
\end{tabular*}
\end{table}

\paragraph{Rank sweeps.}
Rank truncation is applied before gauge alignment.
Table~4 shows representative OLMoE points and Fig.~3 the full cross-model curves.
We treat the sweep as structural evidence rather than a final benchmark.

\end{document}